# IS SHAFER GENERAL BAYES ?[1]

Paul K. Black

Department of Statistics, Carnegie-Mellon University

Pittsburgh, PA 15213

## Abstract

This paper examines the relationship between Shafer's belief functions and convex sets of probability distributions. Kyburg's (1986) result showed that belief function models form a subset of the class of closed convex probability distributions. This paper emphasizes the importance of Kyburg's result by looking at simple examples involving Bernoulli trials. Furthermore, it is shown that many convex sets of probability distributions generate the same belief function in the sense that they support the same lower and upper values. This has implications for a decision theoretic extension. Dempster's rule of combination is also compared with Bayes' rule of conditioning.

## 1. Introduction

Artificial Intelligence projects have employed numerous representations of uncertainty, many of which have been essentially *ad hoc*. However, there has been a general trend towards putting the chosen model of uncertainty on a more theoretical footing. There is still a reluctance to use a full probabilistic representation. This is largely because it is perceived as computationally infeasible and epistemologically inadequate as a representation of uncertainty, though there are some who prefer this approach (Pearl, 1986, Henrion, 1987, and Lauritzen, and Spiegelhalter, 1987). Recently much attention has been focussed on Shafer's *mathematical theory of evidence* (1976); Dempster (1966, 1967) used interval probabilities analagous to Shafer's functions.

This paper addresses the relationship between Dempster/Shafer theory (DS theory) and more traditional views of probability theory, considering both point-valued and interval valued approaches. The comparisons fall into two categories which may be labelled *static* and *dynamic*. At the static level assessment and construction of belief functions are considered. The dynamics concern updating or changing the uncertainty functions. In traditional probability theory this is achieved via Bayes' rule of conditioning. In DS theory Dempster's rule of combination is advocated.

Examples are given showing the relationship between DS functions and closed convex sets of probability distributions. Not all closed convex sets of probability distributions are representable in DS theory. Hence, DS functions must be considered as a subclass of the class of closed convex sets of probability functions. A geometric interpretation of these examples is also given. The examples also demonstrate that a DS model can be generated by many different sets of probability distributions. In particular this has implications for a decision theoretic extension to the theory. Further examples demonstrate difficulties associated with Dempster's rule of combination.

---

[1]This work was supported in part by grant IST-8603493 from the National Science Foundation

## 2. Statics

DS theory considers a *frame of discernment*, which is a field of disjoint events. The theory represents degrees of belief in subsets of the field by three functions named belief, plausibility, and (basic probability) mass, labelled Bel(·), Pl(·) and m(·). The first two correspond to lower and upper values respectively and are derived from m(·) (Shafer, 1976). There is a bijective mapping between Bel(·) and Pl(·). In the special case where these are identical the basic probability function takes values on subsets of unit cardinality only, and hence is a point-valued probability distribution. In this sense DS theory can be said to generalize Bayesian theory. However, DS theory allows an interval representation of uncertainty and may be compared to other theories of interval probability.

Theories of interval probability can usually be equated with convex sets of probability distributions. Let $P$ be a specific set of probability distributions defined on a finite, discrete space given by $\Omega$ and corresponding algebra $A$. Then $P$ is said to be convex if, given any two probability functions $P_1, P_2 \in P$, and $a \in [0,1]$ then:

$$aP_1 + (1 - a)P_2 \in P \tag{2.1}$$

Belief and plausibility functions form a closed convex set in a unit hypercube. This is analagous to particular lower and upper probability functions defining a convex set of probability distributions. The inclusion of the basic probability mass function in the DS model induces a constraint on the space of allowable convex sets of probability distributions. This constraint is the third condition in Shafer's theorem 2.1 (Shafer, 1976, pp. 39) and is generally given as:

$$Bel(A_1 \cup \ldots \cup A_n) \geq \sum_{\substack{K \models \{1,\ldots,n\} \\ K \neq \phi}} (-1)^{|K|+1} Bel(\underset{i \in K}{\&} A_i) \tag{2.2}$$

where '$\models$' represents containment and '&' represents intersection. As Shafer mentions (pp. 35), Choquet (1953) studied such functions extensively in the context of Newtonian capacities. Choquet termed these functions monotone of order p if the above inequality holds for all integers $n \leq p$, and monotone of order infinity if it is monotone of order p for each $p \geq 1$. For two subsets of the frame of discernment the formula reduces to the following:

$$Bel(A_1 \cup A_2) \geq Bel(A_1) + Bel(A_2) - Bel(A_1 \& A_2) \tag{2.3}$$

Other approaches to interval probability theory impose different further constraints on the lower and upper probability distributions. For instance Smith (1961) defines lower and upper conditional probabilities, as well as the marginals, in terms of willingness to take bets.

Shafer appears to regard the basic probability mass functions as fundamental to his theory, though introspection seems to be difficult here and the canonical examples are difficult to follow (but see Shafer, 1981). The basic probability mass function induces a unique belief function, and is recoverable from that belief function. Kyburg (1986) showed that closed convex sets of probability distributions provide a representation of uncertainty that includes Shafer's formulation as a special case. In particular any closed convex set of probability functions can be represented in DS theory subject to the condition given in (2.2). An example demonstrating this result can be found in Dempster (1967) and is summarized in table 1. It can be likened to the throw of a six sided die in which the three possible ways the face up and face down sides sum to seven are each assigned belief of zero and plausibility of 0.5. (This in effect says the probability of a 2, 3, 4 or 5 showing up is at least as great as that of a 1 or 6 showing up etc.) Figure 1 shows the construction of the convex set, introducing the lower and upper values on each marginal consecutively. (The diagrams are projected into two dimensions under the constraint that probabilities sum to one.) The marginals on each of the three possible events determine the boundaries of the convex set (shaded area). It may be reasonable to ask for a geometric interpretation of the convex sets that do not support DS functions.

Table 1: Dempster's example

| Event | 1 | 2 | 3 | 12 | 13 | 23 | 123 |
|---|---|---|---|---|---|---|---|
| Bel(·) | 0 | 0 | 0 | .5 | .5 | .5 | 1 |
| Pl(·) | .5 | .5 | .5 | 1 | 1 | 1 | 1 |
| m(·) | 0 | 0 | 0 | .5 | .5 | .5 | -.5 |

Figure 1: Construction of Dempster's example

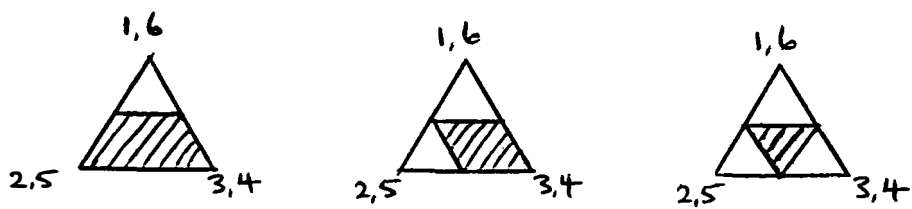

Kyburg (1986) gives an example (attributed to Seidenfeld) based on a mixture model of Bernoulli events that demonstrates his result, and a further example is given by Eddy (1986). Eddy's example can also be related to Bernoulli events. Both of these examples use a frame consisting of four disjoint events and violate Choquet's condition of monotone order 2. Dempster's example uses a frame of three disjoint events and violates the monotone order 3 condition.

The following two examples help to show the serious nature of the problem highlighted by Kyburg.

A. Flip a coin twice, where the coin is known to be biased for heads ($\theta \geq 0.5$).

B. Flip two coins that are known to be biased for heads ($\theta_1, \theta_2 \geq 0.5$, and $\theta_1$ not necessarily equal to $\theta_2$).

Experiment A involves independence and identical distributions. Experiment B merely preserves the independence. Tables 2 and 3 give a full description of the possible outcomes of these two experiments in terms of convex sets and DS functions. The labels H and T refer to obtaining a head or a tail respectively. Notice that the frame of discernment consists of the atoms labelled 1 through 4, and these are necessarily disjoint as required in the DS formulation. The first row simply labels the composite events in a simpler format.

The main point of these examples is that experiment A has a representation in DS theory whereas experiment B does not. Experiment B induces negative values in the basic probability function for the composite elements (124) and (134). This is alarming in that the experiment used as an example here is remarkably simple, (furthermore it is not difficult to demonstrate that the same embarrassment occurs for $\theta_i$ lying in identical proper subintervals of [0,1]. Condition (2.2) is violated in experiment B (i.e. Choquet's monotone order two condition). In particular let $A_1$ be

**Table 2:** Two flips of the same coin, not biased in favour of tails.

$$S = \{ HH, HT, TH, TT \} = \{ 1, 2, 3, 4 \}$$

$$P_\theta = \{ \theta^2, \theta(1 - \theta), (1 - \theta)\theta, (1 - \theta)^2 ; \ 0.5 \leq \theta \leq 1 \}$$

| Event | 1 | 2 | 3 | 4 | 12 | 13 | 14 | 23 | 24 | 34 | 123 | 124 | 134 | 234 | S |
|---|---|---|---|---|---|---|---|---|---|---|---|---|---|---|---|
| Bel(·) | .25 | 0 | 0 | 0 | .5 | .5 | .5 | 0 | 0 | 0 | .75 | .75 | .75 | 0 | 1 |
| Pl(·) | 1 | .25 | .25 | .25 | 1 | 1 | 1 | .5 | .5 | .5 | 1 | 1 | 1 | .75 | 1 |
| m(·) | .25 | 0 | 0 | 0 | .25 | .25 | .25 | 0 | 0 | 0 | 0 | 0 | 0 | 0 | 0 |

**Table 3:** Two flips of different coins, neither biased in favour of tails.

$$S = \{ HH, HT, TH, TT \} = \{ 1, 2, 3, 4 \}$$

$$P_{\theta_1,\theta_2} = \{ \theta_1\theta_2, \theta_1(1 - \theta_2), (1 - \theta_1)\theta_2, (1 - \theta_1)(1 - \theta_2) ; \ 0.5 \leq \theta_1,\theta_2 \leq 1 \}$$

| Event | 1 | 2 | 3 | 4 | 12 | 13 | 14 | 23 | 24 | 34 | 123 | 124 | 134 | 234 | S |
|---|---|---|---|---|---|---|---|---|---|---|---|---|---|---|---|
| Bel(·) | .25 | 0 | 0 | 0 | .5 | .5 | .5 | 0 | 0 | 0 | .75 | .5 | .5 | 0 | 1 |
| Pl(·) | 1 | .5 | .5 | .25 | 1 | 1 | 1 | .5 | .5 | .5 | 1 | 1 | 1 | .75 | 1 |
| m(·) | .25 | 0 | 0 | 0 | .25 | .25 | .25 | 0 | 0 | 0 | 0 | -.25 | -.25 | 0 | (.5) |

(12) and let $A_2$ be (14):

Bel(12 ∪ 14) ≥ Bel(12) + Bel(14) - Bel(12 & 14)

Bel(124) ≥ Bel(12) + Bel(14) - Bel(1)

$$0.5 \geq 0.5 + 0.5 - 0.25 \qquad (2.4)$$

A geometric interpretation of the coin flip examples may give more insight. The convex sets discussed above are four dimensional, this is obviously difficult to picture, but the four dimensions can be projected into three under the constraint of unit total probability. A convenient representation is given by a unit tetrahedron (the coordinates are referred to as *barycentric*).

Let $P$ represent the class of sets of probability distributions, and let $C$ be the class of convex closures of the sets in $P$. If $P \epsilon P$, and $C \epsilon C$ is induced by P according to (2.1), then clearly P is a subset of C. Now let $L$ represent the class of largest convex sets generated by sets in $C$ in the sense that they support the same upper and lower probabilities. If $C \epsilon C$ and $L \epsilon L$ is induced by C then clearly C is a subset of L. If $D$ denotes the set of DS models, (i.e. sets of probability distributions allowable in DS theory) then $D$ is a proper subset of $L$. This is essentially the result that Kyburg (1986) shows, i.e. that the class $L$ is richer than the class $D$. In particular Kyburg shows that there exists $P \epsilon P$ which generates an $L \epsilon L$ (via some $C \epsilon C$) such that L ∉ $D$.

It should be noted that belief function models do not have to be generated from sets of probability distributions. However, all belief functions determine convex sets of probability distributions, and these convex sets can always be generated by some smaller convex sets of probability distributions as described above.

For case A described above (two flips of the same coin) the set $P_\theta \epsilon P$ falls on a quadratic curve between the barycenter and the vertex corresponding to obtaining two heads for certain. $C_\theta \epsilon C$ generated from $P_\theta$ completes the plane in two dimensions. The set $L_\theta$ is then generated by planes supporting the upper probabilities of each of the events 2, 3 and 4. Three more extreme points are created at (0.75, 0.25, 0, 0), (0.75, 0, 0.25, 0) and (0.75, 0, 0, 0.25). The picture for the two different coins is too complicated to describe here, but the following example is instructive.

Example C uses a mixture model of Bernoulli events. Consider an experiment that consists of either flipping a fair coin twice or flipping a two headed coin twice. The parts are performed in some unknown ratio. The convex set of probability functions corresponding to this experiment is given in table 4.

Table 4: Mixture model; flip a fair coin twice, or a two headed coin twice.

$$S = \{ HH, HT, TH, TT \} = \{ 1, 2, 3, 4 \}$$

$$P_\gamma = \{ \frac{1 + 3\gamma}{4}, \frac{1-\gamma}{4}, \frac{1-\gamma}{4}, \frac{1-\gamma}{4} ; 0 \leq \gamma \leq 1 \}$$

| Event | 1 | 2 | 3 | 4 | 12 | 13 | 14 | 23 | 24 | 34 | 123 | 124 | 134 | 234 | S |
|---|---|---|---|---|---|---|---|---|---|---|---|---|---|---|---|
| Bel(·) | .25 | 0 | 0 | 0 | .5 | .5 | .5 | 0 | 0 | 0 | .75 | .75 | .75 | 0 | 1 |
| Pl(·) | 1 | .25 | .25 | .25 | 1 | 1 | 1 | .5 | .5 | .5 | 1 | 1 | 1 | .75 | 1 |
| m(·) | .25 | 0 | 0 | 0 | .25 | .25 | .25 | 0 | 0 | 0 | 0 | 0 | 0 | 0 | 0 |

This is a different convex set to the one described by table 2 yet both convex sets generate exactly the same DS functions. This is because they have the same supporting lines for the lower and upper probabilities. The set of probability functions $P_\gamma$ lie on a straight line that is the line that forms the extremity of the closure of $P_\theta$ in the previous example. $P_\gamma$ and $P_\theta$ share two common distributions, namely the barycenter and the vertex corresponding to two heads for certain. That is, $P_\gamma \neq P_\theta$, and neither is a subset of the other. However $C_\gamma$ is a proper subset of $C_\theta$, and $L_\gamma = L_\theta$. Hence, DS theory is unable to distinguish between the two different convex sets of probability distributions.

A couple of points emerge from this discussion. The first is that a belief function can be generated by many different convex sets of probability distributions. This has consequences for a decision theory based on expected utility arguments. The second is that there are convex sets of probability distributions that cannot be modelled using belief functions, and some of these convex sets are characterized by simple Bernoulli experiments. The examples used above involve different independence relations between the two coin flips. In example A, the flips were assumed to be independent and identically distributed. In B only the independence was maintained, while in C the flips were dependent. The question then remains, which convex sets of probability functions do lend themselves to representation by Shafer functions? Satisfaction of Choquet's condition is a clear requirement which may be related to tacit independence and identical distribution assumptions.

### 3. Implications for Decision Making

The closure of the convex set of probability distributions for example A contains the closure of the convex set for example C and they have equivalent belief function representations. A result in Levi (1980) (and see Seidenfeld, Schervish and Kadane (1987)), shows that it is possible to reach a different decision using expected utility arguments on the smaller set.

Consider the frame of discernment (S) given in examples A and C. Define a gamble as the set of utilities corresponding to events in the frame; for example set utiles for the obtainment of each of the four events in $S$ to be $R = \langle_{g(e)}, 1, 1, -2\rangle$. A decision rule may be stated with respect to a zero option or status quo, $Q$. A gamble $R$ is preferred to $Q$ if the expected utility for that gamble is positive (the expected utility for $Q$ is zero). Suppose a convex set of probability distributions is indexed by $\theta$. Define the set of favourable probability distributions, $P_{R,\theta}$, as those which show preference of gamble $R$ over the zero option.

The expected utility is negative everywhere on the convex set corresponding to the mixture model of example C ( $= -\epsilon(1 + 3\theta)/4$ ); that is, $R$ is dispreferred to the zero option for all probability distributions in that convex set ($P_{R,\gamma}$ is empty). However, the expected utility for the convex set corresponding to example A, $P_\theta$, is positive for some probability distributions provided $\epsilon < 0.5$. That is, $R$ is preferred to the zero option for some of the probability distributions in $P_\theta$ ($P_{R,\theta}$ is not empty). But these two convex sets of probability distributions generate the same DS functions. Hence, based on expected utility arguments, the Dempster/Shafer formulation is not able to distinguish among sets that carry different decisions.

### 4. Dynamics

Dynamics concerns updating or changing opinion in the face of new evidence. The best known rule for updating is Bayes' rule of conditioning. Traditionally this rule applies to point-valued probabilities though Smith (1961) has looked at it's application to interval valued probabilities. DS theory uses Dempster's rule of combination of evidence. Bayes' rule is a rule of conditioning which is suggestive of a temporal quality, i.e. prior evidence is conditioned on new evidence to yield posterior evidence. Dempster's rule is a rule of combination of evidence which is suggestive of an atemporal quality, that is it doesn't matter what order the evidence arrives in, it's combined anyway. This seems to be an important philosophical difference which has not yet been widely considered.

The purpose of this section is to discuss Dempster's rule and give a comparison to Bayes' rule. Under a finite, discrete frame of reference Dempster's rule can give the same results as Bayes' rule (when we are dealing with point-valued probabilities). In this sense Dempster's rule may be regarded as a generalization of Bayes' rule.

Dempster's rule of combination enables computation of the *orthogonal sum* (as Shafer calls it) of two belief functions.

$$m_{12}(A) = \frac{\Sigma_{S\&T = A}\, m_1(S)\, m_2(T)}{1 - \Sigma_{S\&T = \phi}\, m_1(S)\, m_2(T)} \qquad (4.1)$$

where '&' represents intersection, and $A \neq \phi$, together with the requirement that for the empty set, $m_{12}(\phi) = 0$. The Belief function $m_{12}(\cdot)$ then contains the combined evidence of $m_1(\cdot)$ and $m_2(\cdot)$.

In the previous section the problem of flipping a biased coin twice was discussed. Suppose a basic probability mass assignment was obtained for a single flip of the coin. Presumably this assignment would be the same for either flip of the coin provided the flips are independent. The frame $S$ in either case is { H , T }, and for the coin in question the basic probability assignment is m(H) = 0.5, m(T) = 0, m(S) = 0.5. This frame can be refined according to Shafer (1976). A refined frame $S'$ is { HH , HT , TH , TT }. Dempster's rule can now be used to combine these two 'independent' pieces of evidence. The resultant basic probability function is not the same as the one found in table 2. In particular m(14) is zero for this model while m(S) is 0.25. This raises the question of what type of independence is assumed when using Dempster's rule.

Further examples concerning combination of two probability distributions highlight some of the difficulties in using this rule of combination. Zadeh (1984) raised the point that Dempster's rule does not handle conflicting evidence too well. If one basic probability mass function has mass 1-$\epsilon$ on event A, and $\epsilon$ on B, and a second function has 1-$\epsilon$ on C but again has $\epsilon$ on B, then Dempster combination of these two functions yields unit basic probability mass on event B (even though neither of the original functions supported B to a great extent). DS theory is not alone in finding difficulty in coping with conflicting evidence.

Dempster combination of two identical probability distributions also gives cause for concern. Suppose P(H) = 0.25 is assessed by two independent experimenters. Dempster's combination of these two distributions yields P(H) = 0.1!. This again raises the question of what is meant by independence in DS theory. In the case where the experimenter's evidences are independent this may be a reasonable result, but when is this likely to happen?. In the more common scenario that the two experimenters are basing their judgement on similar evidence this result cannot be acceptable.

Shafer (1976, pps. 66/7) exhibits a particular instance where Dempster's rule produces the same numerical results as Bayes' rule. The first requirement is that the "prior" belief function should be a probability function, label this $Bel_1(\cdot)$. Shafer then requires that the "conditioning" function, $Bel_2(\cdot)$, take the form $m_2(A) = 1$ for the new evidence A that is known for certain.

For example, suppose the frame of discernment consists of three disjoint atoms {A, B, C}, with $Bel_1(\cdot)$ defined by $Bel_1(\{A\}) = 0.5$, $Bel_1(\{B\}) = 0.25 = Bel_1(\{C\})$. Now suppose new evidence suggests that either A or B are known for certain, $m_2(\{A,B\}) = 1$. Bayes' rule essentially redistributes the basic probability mass, $Bel_1(\{C\})$, such that the posterior, $Bel_3(\cdot)$ is given by $Bel_3(\{A\}) = 2/3$ and $Bel_3(\{B\}) = 1/3$. It is true that a belief function that assigns unit basic probability mass to the composite event {A,B} does, on Dempster combination with $Bel_1(\cdot)$, yield the same numerical values. However, this is not the only "conditioning" function that will yield the same results. Any belief function that assigns equal mass to {A} and to {B}, and puts the rest of the basic probability mass on {A,B} will, when combined with $Bel_1(\cdot)$, yield the Bayesian result. Further investigation shows that the important requirement for the "conditioning" function is that it satisfies equality for the basic probability mass on all events of the same cardinality. This, in effect, maintains the same ratios in the prior distribution. That summarizes the instances in which Bayes' rule and Dempster's rule will yield the same numerical results without additional assumptions.

## 5. Summary

DS theory may be regarded as a generalization of Bayesian theory in that point-valued probability distributions are representable, and Dempster's rule of combination does produce the same numerical results as Bayes' rule in certain cases. However, DS functions form a subclass of the class of closed convex sets of probability functions. This subclass is caused by satisfaction of Choquet's condition of monotonicity of order infinity. Some convex sets of probability distributions that are not representable in DS theory can be characterized by simple Bernoulli experiments. Choquet's condition may be related to tacit independence assumptions. Furthermore, many convex sets of probability distributions generate the same DS functions in the sense that they support the same lower and upper probabilities. This has implications for a decision theoretic extension involving expected utility arguments. Specific assumptions of dependence for application of Dempster's rule of combination are not clearly defined.

DS theory is making an impact in artificial intelligence, and in particular in rule-based expert systems. However, the theory is computationally exponentially worse off than a traditional probabilistic approach (because DS functions take values on all subsets of the frame of discernment, whereas only the singletons require function values in probability theory). Furthermore the theory cannot model some simple Bernoulli experiments, and cannot coherently use a decision theoretic extension based on expected utility arguments.